\documentclass[wcp]{jmlr}

\usepackage{graphicx}
\usepackage{amsfonts}
\usepackage{amsmath}
\usepackage{amssymb}
\usepackage{bbm}
\usepackage{cleveref}
\usepackage{booktabs}
\usepackage[misc]{ifsym}
\usepackage{comment}
\usepackage{paralist} % inparaenum

\usepackage{pgfplots}

\usepgfplotslibrary{groupplots}
\usetikzlibrary{matrix, positioning}
\pgfplotsset{compat=1.17,
    jitter/.style={
        x filter/.code={\pgfmathparse{\pgfmathresult+(rnd-0.5)*#1}}
    },
    jitter/.default=0.2
}

\usepackage[nolist]{acronym}
\begin{acronym}[UML]
    \acro{KG}{Knowledge Graph}
    \acro{KGE}{Knowledge Graph Embedding}
\end{acronym}

\newcommand{\triple}[3]{\ensuremath{(#1, #2, #3)}}

\newcommand{\entities}{\ensuremath{\mathcal{E}}\xspace}
\newcommand{\relations}{\ensuremath{\mathcal{R}}\xspace}

\newcommand{\kg}{\ensuremath{\mathcal{G}}\xspace}
\newcommand{\scoreFunc}{\phi}
\newcommand{\pair}[2]{(\texttt{#1}, \texttt{#2})}

\usepackage{longtable}% for long tables

\usepackage{booktabs}
\makeatletter
\let\Ginclude@graphics\@org@Ginclude@graphics 
\makeatother

\jmlrvolume{}
\jmlryear{}
\jmlrworkshop{ACML}

\title[Performance Evaluation of KGE Models]{Performance Evaluation of Knowledge Graph Embedding Approaches under Non-adversarial Attacks}

  \author{\Name{Sourabh Kapoor} \Email{
sourabhk@mail.uni-paderborn.de}\\
\Name{Arnab Sharma} \Email{arnab.sharma@upb.de}\\
 \Name{Michael Röder} \Email{
michael.roeder@upb.de}\\
  \Name{Caglar Demir} \Email{
caglar.demir@upb.de}\\
   \Name{Axel-Cyrille Ngonga Ngomo} \Email{
axel.ngonga@upb.de}\\
   \addr Data Science Research Group, Paderborn University}
\begin{document}

\maketitle

\begin{abstract}
\acp{KGE} transform a discrete \ac{KG} into a continuous vector space facilitating its use in various AI-driven applications like Semantic Search, Question Answering, or Recommenders. 
While \ac{KGE} approaches are effective in these applications, most existing approaches assume that all information in the given \ac{KG} is correct.
This enables attackers to influence the output of these approaches, e.g., by perturbing the input.
%However, especially in real-world open \acp{KG}, errors exist.
Consequently, the robustness of such \ac{KGE} approaches has to be addressed.
Recent work focused on adversarial attacks.
However, non-adversarial attacks on all attack surfaces of these approaches have not been thoroughly examined. % where robustness can be understood as a good generalized performance despite perturbations. 
We close this gap by evaluating the impact of non-adversarial attacks on the performance of 5 state-of-the-art \ac{KGE} algorithms on 5 datasets with respect to attacks on 3 attack surfaces---graph, parameter, and label perturbation.
Our evaluation results suggest that label perturbation has a strong effect on the \ac{KGE} performance, followed by parameter perturbation with a moderate and graph with a low effect.
\end{abstract}

%\acresetall
\section{Introduction}

A \ac{KG} is a structured representation of knowledge, typically organized as a multi-relational directed graph where nodes represent entities or concepts, and edges represent relationships between them. The knowledge of real-world facts is represented therein in the form of triples denoted as $\triple{h}{r}{t}$ where $h$ and $t$ correspond to the head and tail entities and $r$ is the relationship between them. %For example, the fact `Paris is the capital of France' can be represented in a KG as $\triple{Paris}{Capitalof}{France}$. 
Due to their effectiveness in representing knowledge, KGs have been used in various areas such as in information retrieval~\citep{DaltonDA14}, question answering~\citep{FerrucciBCFGKLMNPSW10}, and others. To make efficient use of the knowledge representation in KGs, knowledge graph embedding (KGE) models~\citep{bordes2013translating,dettmers2018convolutional} are introduced which aim to capture the complex relationships between entities and relations in KGs. This is done by embedding symbolic representations of KGs into continuous vector spaces by preserving their inherent structure.  

The demand to develop effective KGE models to be applied in various downstream application tasks is ever increasing and that has led to building KGs, harnessing data from public sources, e.g., DBpedia~\citep{AuerBKLCI07}. Although this has led to the benefit of high-quality KGE models to be used for downstream tasks, this has also opened a new {\em attack window} for malicious users. More specifically, the usage of KGE models by utilizing open source KGs as the basis introduces malicious attempts to poison the KGs and thereby the KGE model as well. In recent years, several researchers, studied different {\em adversarial} attack strategies on KGE models by poisoning the KGs or by performing adversarial manipulations of the embedding model~\citep{ZhangZGMSL019,PezeshkpourTS19,BhardwajKCO20,BhardwajKCO21,YouSDZPYF23}. 
The fundamental concept behind these attacks is to focus on a particular fact and manipulate the KGE model to either increase or decrease its {\em plausibility} score. This score represents the likelihood of the fact being true: a higher score indicates a higher probability, while a lower score indicates a lower probability. %The adversarial attacks are preformed by considering a min-max optimization function that attempts to minimize the inclusion or deletion of adversarial or existing triples from the underlying KG. At the same time, the maximum part of the objective function attempts to maximize the objective function of the attacker which is either increasing or decreasing the plausibility of a targeted fact being true. 
Apart from these works that perform targeted adversarial attacks, an attacker might simply perform non-targeted adversarial (or non-adversarial) attacks on the KGE models. %In the machine learning domain, such attacks are prevalent where the attacker simply aims to degrade the performance of the target model by introducing noises, i.e., errors in the training data~\cite{abs-2112-00639}. 
Note that such studies have been carried out for machine learning (ML) models by~\cite{hendrycks2018benchmarking}, but not for the KGE models. These models are frequently used in many critical areas in the web domain. Since the web is a critical point of any country's information sources, an attacker might attempt to disrupt the performance of some critical services (e.g., knowledge-graph-based chatbots on government webpages), thereby destabilizing the country. This kind of attack does not need to have a concrete target and can simply be an attack to degrade the performance of the critical information sources. We can think of such attacks as being similar to denial of service (DoS) attacks. Studying the security of KGE models is only infancy, and some limited works so far have focused on adversarial attacks. However, we do strongly believe that non-adversarial attacks need to be studied to make KGE models that are robust and trustworthy. Therefore, in this work, we study non-adversarial attacks on the KGE approaches considering different attack surfaces. 

In this work, we perform such non-adversarial attacks considering the entire learning framework of the KGE approaches, i.e., performing attacks on the
\begin{inparaenum}
\item knowledge graph, \item parameters, and \item output labels.
\end{inparaenum}
In case of (1), we attack by perturbing the existing triples selected randomly from the KG. More specifically, $k$ percentage of the triples are chosen randomly, and then for each of the selected triples, based on the random value from a Bernoulli distribution, either the head or the relation of the triple is changed (i.e., replaced by some other entity or relation in the same KG).  In (2), the embedding space of the underlying KGE model is targeted where the embedding vectors are perturbed. Herein again a $k$ percentage of the embeddings is selected and then for each of the selected embeddings, either the head or the relation is chosen (based on the Bernoulli distribution). Using a probability distribution, continuous noise is then added to either the head or the relation. Finally, in (3), the labels of randomly selected triples, which in the case of the KGE models are typically the tail entity, are simply flipped, the 0s to 1s and vice-versa.  
We aim to find out the robustness of the existing state-of-the-art KGE models when these non-adversarial attacks in these three levels are done. Precisely, we want to investigate if some KGE models can perform better than others and if so, in which cases and how much it might depend on the underlying KGs. To this end, we have considered 5 datasets and 5 state-of-the-art KGE algorithms to perform non-adversarial attacks. Our results suggest that the attack by performing the label perturbation causes the worst degradation of the performance of the KGE models, followed by parameter and graph perturbations. Moreover, in graph perturbation, for some models, which do not perform well, initially, perturbations can act as a regularizer, thereby improving their overall performance. 

Preliminaries and some formalizations that are used throughout the paper are given in Section~\ref{section:background}. Section \ref{sec:methodology} describes the three different attack approaches that are considered in this work. Section \ref{sec:evaluation} shows details about the experiments and the computational results. We discuss related studies in Section \ref{sec:related-work}.

%Despite having the applications of knowledge graphs in critical areas, the robustness of the knowledge graph embedding (KGE) models is not studied yet extensively. There exist some works such as~\cite{BhardwajKCO20,ZhangZGMSL019,PezeshkpourTS19}, however focus on the robustness considering a specific aspect. More specifically, these works consider investigating the {\em adversarial} robustness by performing {\em poisoning} attacks on the KG, leading to the generation of the {\em poisoned} KGE models which ultimately results in the {\em manipulated} predictions in the downstream tasks. For instance, the works of~\cite{PezeshkpourTS19} proposed gradient based approaches to minimally change the KG in order to maximize the probability that a target fact is learned by the KGE model. Such changes are performed by removing or inserting specific triples in the KG and termed as {\em adversarial graph edits}. Later,~\cite{BhardwajKCO20} rendered the idea of {\em instance attribution} from the {\em interpretable} machine learning domain to identify instances (i.e., triples) which would considerably change the KGE models predictions. In a more recent work~\cite{YouSDZPYF23} proposed to include or remove triples to attack the KGE models by preserving the {\em semantic constraints}.  

\section{Related Work}\label{sec:related-work}

%Most \ac{KGE} models learn continuous vector representations tailored towards link prediction~\cite{chami2020low,hogan2021knowledge}.
%They are often defined as parameterized scoring functions $\scoreFunc_\Theta: \entities \times \relations \times \entities \mapsto \mathbb{R}$,
%where $\Theta$ denotes parameters and often comprise 
%entity embeddings $\mathbf{E} \in \mathbb{V}^{|\mathcal{E}| \times d_e}$, relation embeddings $\mathbf{R} \in \mathbb{V}^{|\mathcal{R}| \times d_r}$, and additional parameters (e.g., affine transformations, batch normalizations, convolutions)~\cite{balavzevic2019hypernetwork,demir2021hyperconvolutional}.
%Since $d_e = d_r$ holds for many models including models reported in~\Cref{table:sota_and_cl}, we will use $d$ to signify the number of real parameters used for the embedding of an entity or 
% CD: resource?
%resource.  
%relation
%Given $\triple{h}{r}{t} \in \entities \times \relations \times \entities$, the prediction $\hat{y}:=\scoreFunc_\Theta\triple{h}{r}{t}$ signals the likelihood of $\triple{h}{r}{t}$ being true~\cite{dettmers2018convolutional,wang2021survey,zamini2022review}.
%Since $\kg$ contains only assertions that are assumed to be true, assertions assumed to be false are often generated by applying the negative sampling, 1vsAll or KvsAll training strategies~\cite{ruffinelli2020you}.
%Throughout this paper, we will denote embeddings with bold fonts, i.e., the embedding of $h$ will be denoted $\mathbf{h}$.
%Moreover, we use $\circ$ and $\cdot$ to denote an element-wise vector multiplication and an inner product in $\mathbb V$, respectively.

In the context of performing malicious attacks on KGE approaches, not many works can be found in the literature, and most importantly, most of them focused on performing adversarial attacks. %such as~\citep{YouSDZPYF23,ZhangZZLXHX23,BhardwajKCO21,BhardwajKCO20,PezeshkpourTS19,ZhangZGMSL019}. %These works proposed methods for {\em attacking} the KGE models by generating adversarial examples. 
For instance, ~\cite{ZhangZGMSL019} first introduced {\em data poisoning} attack strategies to perform adversarial attacks on the KGE models by adding or deleting specific triples. To this end, their strategies follow a two-step process, (a) shifting the embedding of either of the head or tail entities of a target triple to maximize the attack goal, and then (b) adding and/or removing triples from the KG which would facilitate in achieving the goal in (a). The aim in this setting is to degrade or promote the plausibility of a specific fact (i.e., the target triple). %in the learned KGE model. %The perturbations in the form of adding or removing specific triples can be directly related to the target triplet, i.e., share the head or tail entity of the target triplet. However, since such direct perturbations can be detected by the {\em data sanity}, to make the attack stealthy, the authors further proposed indirect attacks that involve perturbing other entities (which are not shared with the entities responsible for the targeted fact) in the KG that would ultimately affect the embedding of the targeted fact.
A later work by~\cite{PezeshkpourTS19} followed a similar sort of setting where they used a gradient-based approach to find out the most influential neighboring triples of the target fact, the removal of which would maximize the attack objective. Searching is performed in the embedding space and then an auto-encoder is used to generate the triples of KG. %However, this approach could only be used for multiplicative KGE models. 
~\cite{BhardwajKCO20} attempted to leverage the inductive capabilities of the KGE models, which are encapsulated by relationship patterns such as symmetry, inversion, and composition within the knowledge graph to perform adversarial attacks. Their approach aims to decrease or increase the model's confidence in predicting target facts by enhancing its confidence in predicting a set of {\em decoy} triples. %More specifically, in the training phase, the aim is to improve the model’s predictive performance on these decoy triples, thereby getting a higher predictive score for the targeted triple through inference properties. 
A further work by them~\citep{BhardwajKCO21} used instance attribution methods from the domain of interpretable ML to perform data poisoning attacks on KGE models. Such attribution methods are first used to identify a (set of) triple(s) in the training set, which contributes most to the prediction of a specific target triple. Then the triple from the training set is either removed or added by replacing one of the two entities of the
influential triple. %with an entity that is far away from the replaced entity in the embedding space. 
\cite {YouSDZPYF23} recently proposed approaches for data poisoning attacks by considering several aspects: black-box attack, poisoning by adding semantically preserving triples, and stealthiness by showing good performance on the {\em cleaned} triples. %To this end, unlike the previous works, they proposed to add {\em indicative paths} containing more than one triple which would maximize the prediction probability of the target poisoned triple being true. 
%None of these works however focused on performing non-adversarial attacks on the KGE approaches. To the best of our knowledge, we are the first in line to study the performance degradation of the KGE approaches under non-adversarial attacks performed at different levels.

Finally, apart from these works which focused on adversarial robustness, there exists a line of works focusing on building KGE models that are {\em robust} to noise in KGs, by~\cite{Xie0LL18,ShanBLJL18,NayyeriVSAYL21,ZhangZZLXHX23}, amongst others. %In other words, these works proposed approaches to develop KGE models performing reasonably well even in the presence of noise in the KG. 
To this end,~\cite{Xie0LL18} first proposed the idea of global and local {\em confidence} scores to identify a tripe as a correct (positive) or a noisy (negative) triple. Assigning scores to triples would help the KGE model to distinguish the correct triple from the noisy ones, thereby dictating the model to learn correctly with the help of the adjusted loss function. %Later the works by~\cite{ShanBLJL18,Cheng2020NoiGANNA,MaZWGHQW22} extended the the idea of confidence score to develop several variants of it. For instance,
~\cite{ShanBLJL18} proposed {\em dissimilarity} measure and {\em support} score alongside confidence score to categorize noisy triples. \cite{Cheng2020NoiGANNA} proposed to use an adversarial training setup, extending the previous works to improve over the works by~\cite{Xie0LL18}. Precisely, they came up with a loss function that makes the KGE models aware of noisy triples. %Precisely, if a triple is noisy, then the plausibility score given by the KGE model would reflect so.  
In a recent work, \cite{ZhangZZLXHX23} proposed a reinforcement learning framework to identify the noisy triples before the training and then remove them. Thus, the KGE model generated in this way would be robust to noise in KG. Note that, all these works consider noise as it is inherently present in KG. Therefore, they proposed approaches to make the KGE models robust against such noise. However, none of them evaluated the performance changes of the KGE models when such noise is added as a form of non-adversarial attacks. %Moreover, all of these works considered noise present in the KG, whereas, in our work, we consider performing non-adversarial attacks on three attack surfaces by inserting noise through perturbation at three different stages in the KGE approaches. To the best of our knowledge, such a kind of study to evaluate the performance degradation of state-of-the-art KGE approaches under such non-adversarial attacks has not yet been done. 

\section{Preliminaries and Notation}
\label{section:background}

Let $\entities$ be the set of entities that are of interest and $\relations$ the set of relations that exist between these entities. 
We express assertions about the entities using triples. A triple $\triple{h}{r}{t}$ comprises a head and a tail entity ($h, t \in \entities$) and a relation $r \in \relations$ that holds between them.
We define a knowledge graph $\kg$ as a collection of triples:%~\cite{dettmers2018convolutional,balavzevic2019tucker,balavzevic2019hypernetwork}:
\begin{equation}
\kg:= \{ \triple{h}{r}{t}\in \entities \times \relations \times \entities\}\,.
\end{equation}
%Thus, a \ac{KG} represents structured collections of assertions describing the world. 

%However, as mentioned by \cite{hogan2021knowledge}, most publicly available \acp{KG} contain missing and erroneous assertions. To this end, a significant amount of work is contributed to inferring triples from an existing set of triples by means of designing logical rules or learning continuous vector representations via knowledge graph embedding models~\cite{nickel2015review}. In this work, we only focus on the latter aspect of inferring triples or the knowledge graph completion process which we formally define next.
%
%\subsection{Knowledge Graph Embeddings}

\acp{KG} are representations of information in a discrete space. However, many modern algorithms cannot process such a graph. Hence, KGE algorithms have been suggested to represent the knowledge of a \ac{KG} in a continuous, low-dimensional embedding space.

Let $\mathbb{V}$ denote a normed-division algebra, e.g. $\mathbb{R},\mathbb{C},\mathbb{H}$, or $\mathbb{O}$~\citep{balavzevic2019hypernetwork,demir2021hyperconvolutional,yang2014embedding,trouillon2016complex,zhang2019quaternion}.
A \ac{KGE} model of a \ac{KG} comprises 
entity embeddings $\mathbf{E} \in \mathbb{V}^{|\mathcal{E}| \times d_e}$ and relation embeddings $\mathbf{R} \in \mathbb{V}^{|\mathcal{R}| \times d_r}$, where $d_e$ and $d_r$ are the size of the embedding vectors. In the following, we use $d$ as size for all embedding vectors, as it has been shown that $d_e = d_r$ holds for many types of models~\citep{nickel2015review}. Throughout this paper, we will denote  embedding vectors with bold fonts, for instance, the embedding of $h$, $r$, and $t$ will be denoted as $\mathbf{h}$, $\mathbf{r}$, and $\mathbf{t}$, respectively.

Given a \ac{KG}, a \ac{KGE} algorithm has the goal to find a \ac{KGE} model that optimizes its scoring function.
Most of these algorithms are tailored towards link prediction~\citep{chami2020low,hogan2021knowledge}, i.e., their scoring function is $\scoreFunc_\Theta: \entities \times \relations \times \entities \mapsto \mathbb{R}$,
where $\Theta$ denotes parameters and often comprise $\mathbf{E}$, $\mathbf{R}$, and additional parameters (e.g., affine transformations, batch normalizations, convolutions). Given an assertion in the form of a triple $\triple{h}{r}{t} \in \entities \times \relations \times \entities$, a prediction $\hat{y}:=\scoreFunc_\Theta\triple{h}{r}{t}$ signals the likelihood of $\triple{h}{r}{t}$ being true~\citep{dettmers2018convolutional}.
Since $\kg$ contains only assertions that are assumed to be true, assertions assumed to be false have to be generated. While different generation methods exist, we will focus on KvsAll~\citep{dettmers2018convolutional}, since recent \ac{KGE} approaches are commonly trained with this strategy~\citep{balavzevic2019hypernetwork,balavzevic2019tucker,nguyen-etal-2018-novel,demir2021convolutional,ruffinelli2020you}. 
%The KvsAll strategy is a multi-label, multi-class classification task, i.e.,

Let $\mathcal{D}=$ denote the training dataset for the KvsAll training strategy. It comprises training data points $(\textbf{x},\textbf{y}) \in \mathcal{D}$ that correspond to unique head entity and relation pairs (\textbf{x}=\pair{h}{r}) that occur in \kg with a binary label vector $\textbf{y} \in \{0,1\}^{|\entities|}$, where
$\textbf{y}_i =1$ for the $i$-th entity $e \in \{ e | \triple{h}{r}{e} \in \mathcal{G} \}$, otherwise $0$.
Consequently, $|\mathcal{D}|$ equals to the number of unique head entity relation pairs in the graph $\mid \{ (h,r) \in \entities \times \relations \mid x \in \entities \land \triple{h}{r}{x} \in \mathcal{G}\}\mid$.
During the training process, most \ac{KGE} algorithms divide the training data into mini-batches. A mini-batch $\mathcal{B}$ consists of $m$ data points with $m \times |\entities|$ binary labels. 
The data points are used to update the entity and relation embeddings $\mathbf{E}$ and $\mathbf{R}$. Hence, during training, a mini-batch is typically represented using the embedding vectors, i.e., $\mathcal{B}$ is expressed as $\pmb{\mathcal{B}} = \{(\pmb{\textbf{x}},\textbf{y})\}$, where $\pmb{\textbf{x}}=(\mathbf{h},\mathbf{r})$ comprises the embedding vectors of $h$ and $r$.
The training is typically performed in several epochs. Within each epoch, all mini-batches are used to update the model's parameters. %including the vectors in $\mathbf{E}$ and $\mathbf{R}$.

%Recent \ac{KGE} models are commonly trained with KvsAll~\cite{balavzevic2019tucker,nguyen-etal-2018-novel,demir2021convolutional,ruffinelli2020you,balavzevic2019hypernetwork}. 
%Using KvsAll can be seen as training a \ac{KGE} model to tackle a multi-label classification problem.

%Moreover, we use $\circ$ and $\cdot$ to denote an element-wise vector multiplication and an inner product in $\mathbb V$, respectively.
%

%
% \paragraph{Optimization:} The binary cross-entropy loss is commonly used with the 1vsAll and KvsAll training strategies, with the incurred loss for a training data point ($\mathbf{x}=\pair{h}{r}, \mathbf{y}$) given by
% \begin{equation}
% \ell(\mathbf{\hat{y}},\mathbf{y}) =\sum\limits_{i=1}^{|\entities|}  -\mathbf{y}^{(i)} \text{log}(\hat{\mathbf{y}}^{(i)}) - \big( 1-\mathbf{y}^{(i)} \big) \text{log}\big( 1-\hat{\mathbf{y}}^{(i)} \big),
% \label{eq:kvsall_binary_cross_entropy_loss}
% \end{equation}
% where $\textbf{y} \in [0,1]^{|\entities|}$ and $\mathbf{\hat{y}}:=\scoreFunc_\mathbf{w}(\mathbf{x}) \in [0,1]^{|\entities|}$.

\section{Methodology}
\label{sec:methodology}
%
%Let $\mathcal{D}= \{ (\textbf{x}_i,\textbf{y}_i) \}_{i=1} ^n$ denote a training dataset for the KvsAll training strategy.
%&A training data point $(\textbf{x},\textbf{y}) \in \mathcal{D}$ corresponds to a \emph{unique} \emph{occurring} head entity and relation pair (\textbf{x}=\pair{h}{r}) with a binary label vector $\textbf{y} \in [0,1]^{|\entities|}$, where
%$\textbf{y}_i =1$ for the $i$-th entity $e \in \{ e | \triple{h}{r}{e} \in \mathcal{G} \}$, otherwise $0$.

In this work, we perform non-adversarial attacks on \ac{KGE} algorithms, i.e., our goal is to reduce the performance of \ac{KGE} models in a downstream task.
To this end, we use three different attack surfaces:
\begin{inparaenum}
    \item the input knowledge graph,
    \item the target labels, and
    \item the model parameters.
\end{inparaenum}
Every attack incorporates a parameter $k$ which allows us to regulate the extent to which data is perturbed during the attack.

%TODO Introduction
%\begin{itemize}
%    \item How do we want to reach our goal?
%    \item What are attack surfaces (in general and for KGE algorithms specifically)? Two attacks on input data and one attack on the parameters of the KGE
%\end{itemize}

% CD : Maybe brief motivation for three type of perturbation
\subsection{Graph Perturbation}
\label{subsec:DiscretePerturbation}

The first attack surface that we look at is the training data that is gathered from the \ac{KG}. During this attack, the attacker perturbs $k\%$ of the input data within each mini-batch by changing the head or relation information of the data points. Let $\mathcal{B}$ be a mini-batch and let $\mathcal{B}^{\star} \subset \mathcal{B}$ be a randomly sampled subset comprising $k\%$ of the training examples of $\mathcal{B}$. The attacker replaces the original mini-batch with a perturbed version of the batch by replacing $\mathcal{B}^{\star}$ with ${\mathcal{B}^{\star}}'$:
\begin{equation}
    \mathcal{B}'= {\mathcal{B}^{\star}}' \cup (\mathcal{B} \backslash \mathcal{B}^{\star})\,,
\end{equation}
with $|{\mathcal{B}^{\star}}'|=|\mathcal{B}^{\star}|$.
Hence, the graph Perturbation (GP) attack is defined as generating ${\mathcal{B}^{\star}}'$ based on $\mathcal{B}^{\star}$ by perturbing the head or relation information that has been gathered from the \ac{KG}. 
Let $(\textbf{x}_i,\textbf{y}_i) \in \mathcal{B}^{\star}$ be the $i$-th data point in $\mathcal{B}^{\star}$. Let $\xi_i$ be the $i$-th random value sampled from a Bernoulli distribution with a probability of $0.5$ being either 0 or 1. Let $h_i'$ and $r_i'$ be randomly sampled elements from $\entities$ and $\relations$, respectively. Within this attack, an attacker perturbs the data point $(\textbf{x}_i,\textbf{y}_i)$ by creating $\textbf{x}_i'$ as follows:
\begin{equation}
    \textbf{x}_i'= 
    \begin{cases}
    (h_i',r_i) & \text{if } \xi_i = 0,\\
    (h_i,r_i') & \text{else}.
    \end{cases}
\end{equation}
This perturbation is applied to all data points in $\mathcal{B}^{\star}$ to form  ${\mathcal{B}^{\star}}'$:
\begin{equation}
    {\mathcal{B}^{\star}}' = \{(\textbf{x}_i', \textbf{y}_i) | (\textbf{x}_i,\textbf{y}_i) \in \mathcal{B}^{\star} \}\,.
\end{equation}

To give an example of such a perturbation let us assume one of the data points from the set $\mathcal{B}^*$ as $\mathbf{x}=\pair{:Einstein}{:bornIn}$. If $\xi_i=0$ then the perturbed point could be $\mathbf{x}'=\pair{:Laplace}{:bornIn}$, whereas if $\xi_i \neq 0$ then $\mathbf{x}'=\pair{:Einstein}{:capitalOf}$.

% \begin{equation}
%     \mathcal{B}'= \{(\mathbf{x}'_i=(h'_i,r'_i), \mathbf{y}_i)\}_{i=1}^{|\mathcal{B}_k|} \cup \{(\mathbf{x}_i=(h_i,r_i), \mathbf{y}_i)\}_{i=1}^{|\mathcal{B} \ \mathcal{B}_k|}
% \end{equation}

%Such perturbation can be conducted before the training by means of adding perturbed data points into $\mathcal{D}$ or by means of perturbing data points in a mini-batch. 

Note that, the idea of manipulating the training data by adding some specific triples in order to make the trained model giving specific predictions is termed as {\em data poisoning} or adversarial attacks~\citep{BhardwajKCO20,ZhangZGMSL019,PezeshkpourTS19}. In the context of knowledge graph embeddings, different attack strategies are gaining popularity due to the critical downstream applications of KGE models. 
The goal of an attacker is to introduce {\em malicious} facts in terms of adding triples in the training data, leading to the generation of {\em poisoned} KGE models.
In this work, however, we do not aim to perform such adversarial attacks on the KGE models. More specifically, we do not have a specific target of introducing a fact, rather, we perform non-adversarial attacks by perturbing either the head or the relations of specific triples in the graph and then finally measure how the perturbation would affect the robustness of the KGE models.
%To this end, to the best of our knowledge, this is the first such a work which considers evaluating such a scenario.   

\subsection{Label Perturbation}

The Label Perturbation (LP) is a similar attack as GP. However, within this attack, the attacker perturbs a data point $(\textbf{x}_i,\textbf{y}_i)$ by inverting the label vector as follows:
\begin{equation}
    \textbf{y}_i'= \{\neg y_{i,j} | y_{i,j} \in \textbf{y}_{i}\}\,.
\end{equation}
This perturbation is applied to all data points in $\mathcal{B}^{\star}$ to form  ${\mathcal{B}^{\star}}'$:
\begin{equation}
    {\mathcal{B}^{\star}}' = \{(\textbf{x}_i, \textbf{y}_i') | (\textbf{x}_i,\textbf{y}_i) \in \mathcal{B}^{\star} \}\,.
\end{equation}
For a data point with $\mathbf{x}_i=\pair{:Einstein}{:bornIn}$ and a vector $\mathbf{y}_i$ filled with zeros except for a single 1 at the id of the entity \texttt{:Ulm}, the attack would perturb the label vector by inverting all its values. Hence, the new label vector $\mathbf{y}_i'$ would express that the embedding model is expected to predict that the entity \texttt{:Einstein} has the relation \texttt{:bornIn} to all entities, except \texttt{:Ulm}.

In ML, label perturbation of training data is commonly employed to mitigate overfitting and noise. For instance,~\cite{SzegedyVISW16} used a variant of label perturbation called label smoothing to improve the generalization performance. There exist other such methods such as bootstrapping loss by~\cite{ReedLASER14} and label correction by~\cite{PatriniRMNQ17}, which are different types of label perturbation techniques, introduced to generate robust models. Some studies have further explored strategies for adversarial attacks on deep learning models via label perturbations~\citep{songmulti,zhangrobust}.  However, none of these prior works considered knowledge graphs in this context, which we look into.
The attack defined above is a direct application of the method proposed by~\cite{songmulti,zhangrobust}.

\begin{comment}
can be seen as adding discrete noise into the learning process by means of flipping 1s or 0s in $\textbf{y}_i$ with a specific probability $\mathbb{P}$. More specifically, to generate a KGE model we get a training dataset $\mathcal{D}$ which is a set of instances in the form $(\mathbf{x},\mathbf{y})$, where $\mathbf{x}$ corresponds to a head entity and relation pair, and $\mathbf{y}$ is the label vector corresponding to $\mathbf{x}$. Similar to the former two perturbations, here also we do it on a set of data points $\mathcal{B}_k$ selected uniformly at random from a mini-batch $\mathbb{B}$. However, in this case for each of the data points $(\mathbf{x},\mathbf{y})$, where $\textbf{y} \in [0,1]^{|\entities|}$ we just flip the values of the elements of $\mathbf{y}$. Let us assume after flipping, we get the following perturbed mini-batch, 

\begin{equation}
    \mathcal{B}'= \{(\mathbf{x}_i, \mathbf{y}'_i)\}_{i=1}^{|\mathcal{B}_k|} \cup \{(\mathbf{x}_i, \mathbf{y}_i)\}_{i=1}^{|\mathcal{B}|-|\mathcal{B}_k|}
\end{equation}

The label perturbation function $LP$ therefore takes a mini-batch $\mathcal{B}$ and a specific probability distribution for flipping the 1s and 0s in $\mathbf{y}_i$ to finally generate a mini-batch $\mathcal{B}'$. This can be defined as follows,

\begin{equation}
     LP: \mathcal{B} \times \mathbb{P} \rightarrow  \mathcal{B}',
\end{equation}
\end{comment}

\subsection{Parameter Perturbation}
\label{subsec:ParameterPerturbation}

The third attack surface does not focus on the training data but on the learned parameters. This Parameter Perturbation (PP) changes $k\%$ of the learned vectors, before each of the training epochs.
More formally, let $\pmb{\mathcal{B}}$ be the vector-based representation of a mini-batch $\mathcal{B}$ between two epochs and let $\pmb{\mathcal{B}}^{\star} \subset \pmb{\mathcal{B}}$ be a randomly sampled subset comprising $k\%$ of the training examples of $\pmb{\mathcal{B}}$. The attacker replaces the original mini-batch with a perturbed version of the batch by replacing $\pmb{\mathcal{B}}^{\star}$ with ${\pmb{\mathcal{B}}^{\star}}'$:
\begin{equation}
    \pmb{\mathcal{B}}'= {\pmb{\mathcal{B}}^{\star}}' \cup (\pmb{\mathcal{B}} \backslash \pmb{\mathcal{B}}^{\star})\,.
\end{equation}
The attack is defined as generating ${\pmb{\mathcal{B}}^{\star}}'$ by perturbing the head or relation vectors in $\pmb{\mathcal{B}}^{\star}$. 
Let $(\pmb{\textbf{x}}_i,\textbf{y}_i) \in \pmb{\mathcal{B}}^{\star}$ be the $i$-th data point in $\pmb{\mathcal{B}}^{\star}$. Let $\xi_i$ be the $i$-th random value sampled from a Bernoulli distribution with a probability of $0.5$ being either 0 or 1. Let $\textbf{q}$ be a $d$-dimensional vector with randomly sampled values.
%Let $h_i'$ and $r_i'$ be randomly sampled elements from $\entities$ and $\relations$, respectively. 
Within this attack, an attacker perturbs the data point $(\pmb{\textbf{x}}_i,\textbf{y}_i)$ by creating $\pmb{\textbf{x}}_i'$ as follows:
\begin{equation}
    \textbf{x}_i'= 
    \begin{cases}
    (\mathbf{h} + \textbf{q},\mathbf{r}) & \text{if } \xi_i = 0,\\
    (\mathbf{h},\mathbf{r} + \textbf{q}) & \text{else}.
    \end{cases}
\end{equation}

Some existing works showed that such perturbations could be used by attackers to attack the learned models. %The idea herein is that the attackers have full access to the parameters of a learned model. Then some specific parameters are selected to perturb to introduce malicious behavior in the model. 
For instance, ~\cite{KuritaMN20} proposed an optimization algorithm to perturb the weights of a DNN model in such a way so that whenever specific feature values are present in the input, the output will be predicted to a specific class.\footnote{Note that, in the literature, such attacks are also called {\em trojan attacks} where the attacker aims that the model predicts a specific class when some specific feature values are present~\citep{LiuMALZW018}.} There is a different line of work by~\cite{BaiWZL0X21} that performs such kind of perturbation on the models' parameters, however on the memory level by flipping the bits of the parameters.
However, similar to the perturbation performed on the graph and labels, the works mentioned above belong to a line of works that perform adversarial attacks with a specific goal in mind. For KGE models no such works have considered evaluating the models' performance when non-adversarial attacks are performed on the learned parameters. 
\section{Evaluation}\label{sec:evaluation}
In this section, we first describe the experimental setup of our evaluation describing the datasets and the models we consider in this work. Next, we report the results of our evaluation based on the three attack surfaces we look into.
\subsection{Datasets}
\label{subsection:datasets}

Table~\ref{table:KGdatasets} lists the datasets and their features that we use for the evaluation of the impact of the non-adverserial attacks on \ac{KGE} algorithms.
%In our study on non-adversarial attacks, we utilized several benchmark datasets, including UMLS, KINSHIP, WN18RR, NELL-995 h100, and FB15k-237. 
The UMLS dataset by~\cite{McCray2003} contains 135 medical entities and their connections using 46 distinct relations. 
The KINSHIP dataset by~\cite{Denham2014TheDO} describes the Alyawarra tribe's kinship dynamics with 25 unique relationship types. 
Apart from these two smaller datasets, we also use three larger datasets.
WN18RR by~\cite{dettmers2018convolutional} is a version of WordNet optimized for the link prediction task proposed by~\cite{Bordes2013WN18RR}.
NELL-995-h100 is a subset of the Never-Ending Language Learning dataset by \cite{xiong-etal-2017-deeppath}.
FB15k-237 by~\cite{toutanova-chen-2015} is a subset of the Freebase knowledge graph.%~\cite{bollacker2008freebase}.
%In case of large scale \acp{KG}, we conduct experiments using three datasets: FB15k-237, a subset of Freebase \cite{bollacker2008freebase}; WN18RR, a version of WordNet optimized for the link prediction task~\cite{Bordes2013WN18RR}; and NELL-995-h100, a version of the Never-Ending Language Learning dataset \cite{xiong-etal-2017-deeppath}.
\begin{table}[htb]
    \caption{Datasets used throughout the evaluation and their features (number of entities, relations, and triples in each split).}
    \centering
    \small
    \label{table:KGdatasets}
    \begin{tabular}{l@{\hspace{3mm}}c@{\hspace{3mm}}c@{\hspace{3mm}}c@{\hspace{3mm}}c@{\hspace{3mm}}c}
    \toprule
    \textbf{Dataset} & $|\mathcal{E}|$&  $|\mathcal{R}|$ & $|\mathcal{G}^{\text{Train}}|$ & $|\mathcal{G}^{\text{Validation}}|$ &  $|\mathcal{G}^{\text{Test}}|$\\
    \midrule
    UMLS~\citep{McCray2003}          &135      &46  &5,216       &652&661\\
    KINSHIP~\citep{Denham2014TheDO}       &104      &25  &8,544       &1,068&1,074\\
    WN18RR~\citep{dettmers2018convolutional}        &  40,943  & 22  &86,835     & 3,034  & 3,134 \\
    NELL-995-h100~\citep{xiong-etal-2017-deeppath}  & 22,411  &43   &50,314     &3,763&3,746\\
    FB15K-237~\citep{toutanova-chen-2015}     &  14,541 & 237 &272,115    & 17,535 & 20,466 \\

    \bottomrule
    \end{tabular}
\end{table}

\subsection{Experimental Setup}

% 1. KGE approaches
% 2. Overall experiment structure
% 3. KGE hyperparameters and other details

Throughout our evaluation, we use 5 \ac{KGE} algorithms with different embedding spaces: DistMult ($\mathbb{R}$)~\citep{yang2014embedding}, ComplEx ($\mathbb{C}$)~\citep{trouillon2016complex}, QMult ($\mathbb{H}$)~\citep{zhang2019quaternion}, MuRE ($\mathbb{R}$)  ~\citep{balažević2019multirelational} , and Keci~\citep{demir2023clifford}.
With our experiments, we compare the performance of these \ac{KGE} algorithms  on the aforementioned datasets with and without non-adversarial attacks.
%To this end, we first evaluate performances of DistMult, ComplEx, QMult, Keci, and MuRE across the aforementioned datasets without any perturbation. 
For each attack described in Section~\ref{sec:methodology}, we evaluate the performance of the algorithms using an increasing perturbation ratio $k \in \{0,$ $0.01,$ $0.02,$ $0.04,$ $0.08,$ $0.16,$ $0.32,$ $0.64\}$. %  across all levels, where k denotes the ratio of data points perturbed in a batch.
For attack surfaces, that rely on probability distributions, we use an even distribution.
We repeat each experiment 5 times with different seed values for random number generators.
We measure the \ac{KGE} performance in terms of Hits@N and Mean Reciprocal Rank (MRR). However, we only report the MRR values on the test data within this paper due to the brevity of this work.\footnote{https://figshare.com/s/4367528fa5c6af381a5a}
For each \ac{KGE} approach, we choose the size of embedding vectors $d$ so that all vectors can be represented as 32-dimensional real-valued vectors. %, as in~\cite{demir2023clifford,chami2020lowdimensional}.
%In our study, each entity and relation across all datasets was represented using 32-dimensional real-valued vectors, adhering to the methodologies described in \cite{demir2023clifford}. 
Furthermore, we apply a consistent set of hyperparameters across all experiments.
We use a learning rate of 0.1, a training duration of 100 epochs, a mini-batch size of 1024, and the KvsAll scoring technique. %~\cite{demir2021convolutional,demir2023clifford,ruffinelli2020you}.
Additionally, for the Keci algorithm, we set its two additional parameters \( p = 0 \) and \( q = 1 \) to the default values suggested by~\cite{demir2022hardware}.
% We also utilize two different distribution for parameter perturbation namely: uniform and Gaussian. where scaler is set to 0.3. % For our evaluation on different perturbation levels, we report the MRR scores of  five different KGE models - ComplEx, DistMult, Keci, Qmult, and MuRE - based on their consistent performance on link prediction tasks.
%Thereafter, we evaluate the performances of the aforementioned models considering different levels of perturbations on the same benchmark datasets. 
%To ensure a fair comparison, we maintain consistent hyperparameters when evaluating the impact of perturbations across attack surfaces and in scenarios without perturbations.

\subsection{Results}

\subsubsection{Graph Perturbation.}

\begin{figure}[htb]
  \centering
  \includegraphics[width=0.8\linewidth]{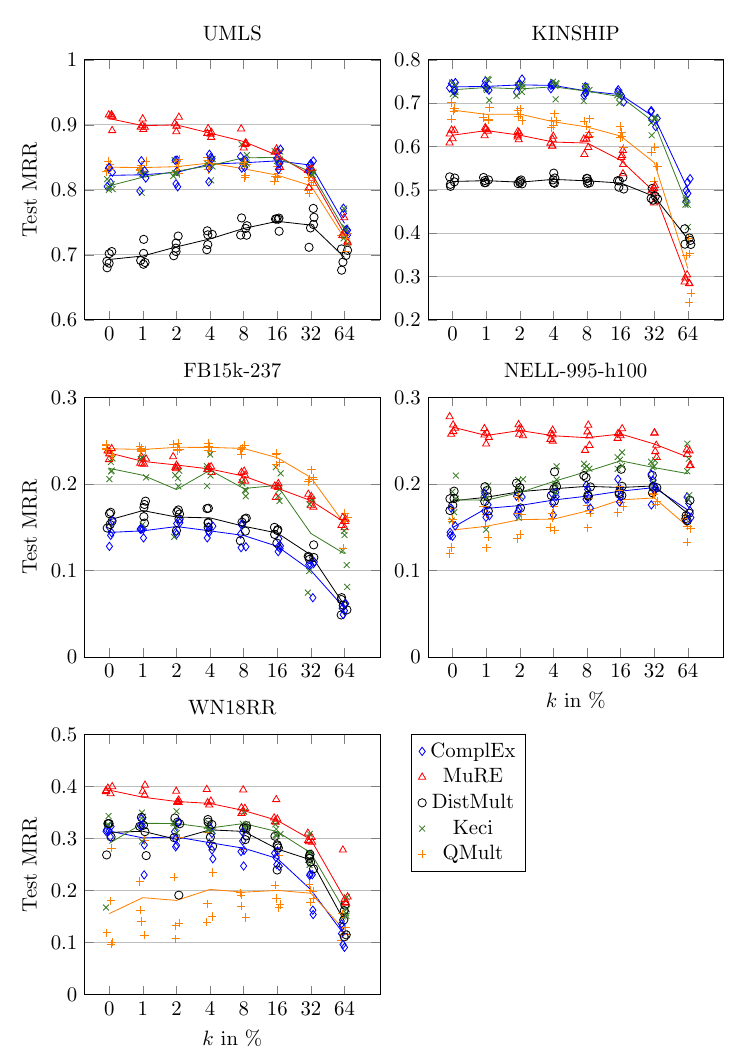}
  \caption{Test MRR performance of the KGE approaches on different datasets with Graph Perturbation and varying perturbation ratios.}
  \label{fig:dp-plots}
\end{figure}

\Cref{fig:dp-plots} reports the average test MRR performances of the aforementioned \ac{KGE} algorithms with different ratios of graph perturbation on the aforementioned datasets.
The results show that on 4 out of 5 datasets, nearly all \ac{KGE} algorithms show a clear decline in MRR at higher perturbation levels, specifically at 32\% and 64\% perturbation ratios.
On the \textit{FB15k-237} and WN18RR datasets, the decline already starts earlier, e.g., MuRE and Keci show significant performance reductions starting from 8\% perturbation on FB15k-237.
Only on the NELL-995-h100 dataset, our results suggest that the perturbations have little to no effect on the test MRR.
In addition, the results on WN18RR and NELL-995-h100 indicate that the QMult \ac{KGE} model is sensitive to the randomness induced which is demonstrated by the varied performance of QMult at same ratios but for different random seeds.

\subsubsection{Label Perturbation.}

\begin{figure}[htb]
  \centering
  \includegraphics[width=0.8\linewidth]{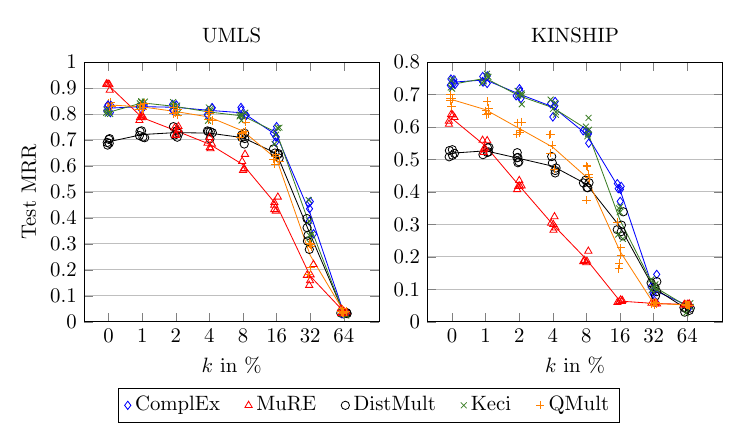}
  \caption{Test MRR performance of the KGE approaches on the UMLS and KINSHIP
datasets with Label Perturbation and varying perturbation ratios.}
  \label{fig:lp-plots}
\end{figure}

The impact of the Label Perturbation on the \ac{KGE} algorithm performance is larger compared to the Graph Perturbation. 
\Cref{fig:lp-plots} shows that the MRR on the test set drops dramatically with higher perturbation rates on the small UMLS and KINSHIP datasets.
On the other three datasets, the effect is even more severe. Even a small perturbation rate of 0.1\% already causes the MRR of all \ac{KGE} approaches to drop close to 0.

\begin{comment}    
In this subsection we quantify %(is it the right word) 
the effect of attack on the label surface on the robustness of the \ac{KGE} algorithm and by far this is the most sensitive and most vulnerable attack surface in the study where even a small perturbation ratio like 0.1\% brings down the performance of all models across large scale datasets to nearly zero. 
This is expected from the nature of the KvsAll training strategy and the definition of label perturbation in section 4, where even perturbing one sample by flipping the labels would add n new incorrect triples in the training dataset disrupting the embedding process significantly. 
However the result is not so pronounced on the smaller datasets like UMLS and KINSHIP due to the smaller. For instance,~\Cref{UMLS_LP} the significant downfall in test MRR could be seen after the perturbation ratio 8\% with an exception to MuRE. This behaviour on smaller datasets can be explained by the less number of unique entities, 135 and 104, in UMLS and KINSHIP respectively. 
\end{comment}

%\input{images/LP}

% \begin{figure}
%     \centering
%     \input{images/UMLS_LP}
%     \caption{Test MRR performance of KGE models with different noise ratios and random seeds on UMLS and label perturbation.}
%     \label{UMLS_LP}
% \end{figure}
% \begin{figure}
%     \centering
%     \input{images/KINSHIP_LP}
%     \caption{Test MRR performance of KGE models with different noise ratios and random seeds on KINSHIP and label attack surface.}
%     \label{KINSHIP_LP}
% \end{figure}

\subsubsection{Parameter Perturbation.}
% In~\Cref{UMLS_PP,KINSHIP_PP,FB15_PP,NELL_PP,WN18RR_PP} this subsection explores the impact of parameter perturbation on \ac{KGE} algorithms across several datasets. Parameter perturbation, unlike graph perturbation, involves direct adjustments to the model's learned parameters, providing insights into model robustness and error tolerance under modified entity and relation embedding vectors during the training. For a fair comparison we utilize the same benchmark datasets and \ac{KGE} models used in the graph perturbation.

\begin{figure}[htb]
  \centering
  \includegraphics[width=0.8\linewidth]{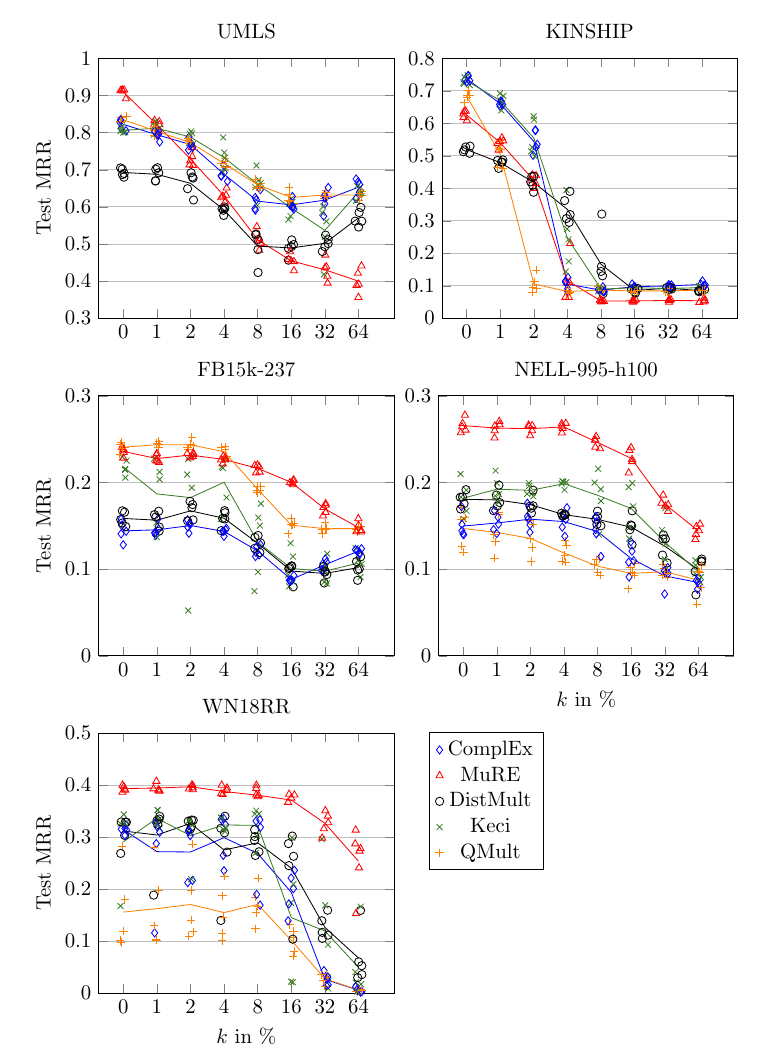}
  \caption{Test MRR performance of the KGE approaches on different datasets with Parameter Perturbation and varying perturbation ratios.}
  \label{fig:pp-plots}
\end{figure}

\Cref{fig:pp-plots} summarizes the results of the Parameter Perturbation experiments. 
On all datasets, the performance of the \ac{KGE} decreases with perturbation rates of 16\% or higher. On the small datasets, the effect already starts with a perturbation rate of 1 or 2\%. In contrast, small perturbation ratios do not seem to have a big influence on the \ac{KGE} algorithm performance when tested with larger datasets.
As in the Graph Perturbation experiment, QMult shows to be sensitive to the seed value of an internal random number generator, which is again demonstrated by the varied performance of QMult on WN18RR.

%\clearpage
% \begin{figure}
%     \centering
%     \input{images/UMLS_PP}
%     \caption{Test MRR performance of KGE models with different noise ratios and random seeds on UMLS and parameter attack surface.}
%     \label{UMLS_PP}
% \end{figure}
% \begin{figure}
%     \centering
%     \input{images/KINSHIP_PP}
%     \caption{Test MRR performance of KGE models with different noise ratios and random seeds on KINSHIP and parameter attack surface.}
%     \label{KINSHIP_PP}
% \end{figure}

% \begin{figure}
%     \centering
%     \input{images/FB15k_PP}
%     \caption{ Test MRR performance of KGE models with different noise ratios and random seeds on FB15k-237 and parameter attack surface. }
%     \label{FB15_PP}
% \end{figure}

% \begin{figure}
%     \centering
%     \input{images/NELL_PP}
%     \caption{Test MRR performance of KGE models with different noise ratios and random seeds on NELL-995-h100 and parameter perturbation.  }
%     \label{NELL_PP}
% \end{figure}

% \begin{figure}
%     \centering
%     \input{images/WN18_PP}
%     \caption{Test MRR performance of KGE models with different noise ratios and random seeds on WN18RR and parameter perturbation. }
%     \label{WN18RR_PP}
% \end{figure}

\subsection{Discussion}

\begin{comment}
\todo[inline]{SENTENCES FROM ABOVE START}
This trend indicates a robustness challenge for models under high perturbation scenarios.
Contrary to this trend, the performance of various \ac{KGE} models indicates increased performance at smaller perturbation ratios in comparison to no perturbation, with an exception to MuRE in UMLS[\ref{UMLS_DP}]. This might suggest that models like DistMult tend to overfit to the training data and the attack at this surface might not be as beneficial as compared to large perturbation efforts, particularly in \textit{UMLS}.

This variation suggests or supports the fact of overfitting of data across models on the NELL dataset.

This is expected from the nature of the KvsAll training strategy and the definition of label perturbation in section 4, where even perturbing one sample by flipping the labels would add n new incorrect triples in the training dataset disrupting the embedding process significantly. 
This behaviour on smaller datasets can be explained by the less number of unique entities, 135 and 104, in UMLS and KINSHIP respectively.

\todo[inline]{SENTENCES FROM ABOVE END}
\end{comment}

The results reported above indicate that all tested \ac{KGE} algorithms can be influenced by perturbing the data on nearly all datasets.
However, the results vary depending on the \ac{KGE} algorithm, attack surface, and dataset size. Below we further discuss them. %based on the %three attack surfaces that we considered.

\subsubsection{Graph Perturbation.}

The results of the Graph Perturbation experiments allow two conclusions. 
First, although a non-adversarial attack on the graph input, i.e., on $\mathbf{x}$, has a negative effect on the performance, the perturbation rate has to be higher than for the other two attack surfaces to achieve a similar reduction.
Second, in some cases, a small perturbation showed the opposite effect, i.e., the performance of some \ac{KGE} algorithms increases, such as DistMult on UMLS and the results of all approaches except MuRE on  NELL-995-h100.
A similar result can be seen for the MRR measured on the validation split of the datasets. %\footnote{See the supplementary material for results on the training and validation splits.}
Hence, we conclude that the perturbed data acts like a regularizer in the training process of some \ac{KGE} algorithms, making them less vulnerable to overfitting and, hence, boosting their performance on the test and validation data. Such behavior is not surprising and there are works by~\cite{OrvietoRKB23} who mention the effect of explicit regularization in the ML model by performing perturbation on the data (e.g., via injecting noise). In this work, we observe a similar effect. %while learning the KGE model parameters. %Therefore, we can use this insight to further develop KGE models which can have improved generalization performance.

\begin{comment}
In the first attack surface, low levels of perturbation can beneficially reduce overfitting—evidenced by minor improvements or stability in Test MRR. Hence low level of attacks on this surface maybe acts as a regularizer to the model performing inferior as compared to the others. The overarching trend indicates that excessive perturbation severely hampers model performance.  Across all datasets, the trend is clear: graph perturbations significantly affect the performance of KGE models, with a more pronounced impact as the perturbation level increases. Models like ComplEx and MuRE generally show higher initial MRR but suffer considerable performance losses at high perturbation levels. % DistMult particularly has slight positive change in performance at lower ratios of perturbation.
The empirical findings from this analysis underline a complex interaction between perturbation, overfitting, and model performance in KGE models. In general if a model is performing well even a small perturbation in the graph hinders its performance.
%For instance MuRE demonstrate a negative impact of perturbation across all datasets 
These trends underscore the nuanced impact of perturbation on the robustness and accuracy of KGE models across different knowledge domains.
\end{comment}

\subsubsection{Label Perturbation.}

The attacks on the label vector $\mathbf{y}$ showed the highest impact in our experiments.
The reason for the high impact can be explained by comparing an attack on a single label vector with the number of edges that would have to be added by a Graph Perturbation attack to achieve a similar effect.
Consider a training example $(\mathbf{x},\mathbf{y})$ in which the label vector has only a single 1 and all other values are 0. 
If this vector is inverted, the attack has an effect of adding $|\entities| - 1$ edges to the graph. For example, for the WN18RR dataset, changing a single label vector would add more than 40k false edges to a graph that contains 86k edges. After changing a single vector, nearly $1/3$ of the training data that the \ac{KGE} algorithms rely on becomes faulty.
This effect is bigger, the larger and the more sparse the graphs are. 
%\clearpage
\noindent
The two smaller graphs UMLS and KINSHIP have a small number of entities and a high node degree with 38.6 and 82.1 edges per node, respectively.
The impact of a small perturbation rate is not as big as on the large datasets. WN18RR, NELL-995-h100 and FB15k-237 have a node degree of 2.1, 2.2, and 18.7, respectively. Changing only 0.1\% of the label vectors already adds 3.5M, 1.1M, and 3.9M faulty triples to the training data of these datasets, which are many more triples than the size of the training split. Therefore, label perturbation has been shown to be much more effective in degrading the performance as it can entirely change the structure of the underlying KG. 
%Since to the best of our knowledge, there does not exist any work in the domain of KGEs that considers such flipping attacks, we simply consider the typical attack methods as described in ML literature by~\cite{songmulti,zhangrobust}.
After considering the results obtained through the experiments we can conclude that the typical label-flipping attacks suggested by ~\cite{songmulti,zhangrobust} for ML algorithms cannot simply be used for KGEs. Future works proposing new techniques pertaining to label-flipping attacks for KGE models need to be studied. Since this would require a significant extension of the current paper we consider it as a possible future work.

%Furthermore, label perturbation that is considered in as a way of improving the performance of the corresponding deep learning models, is fundamentally different compared to our approach. More specifically, in the former cases, some specific data points are chosen, and for each of the points, a single label is changed. On contrary, for KGE approaches when we perform the label perturbation, the values of each chosen label vector are flipped, which do not anymore work as a regularizer and degrades the performance of the underlying model significantly.

\subsubsection{Parameter Perturbation.}

Attacks on the parameter surface show a stronger negative impact on the performance of \ac{KGE} algorithms when compared to the Graph Perturbation attack. When compared to the Label Perturbation attack, the dataset size seems to have a large influence. On larger datasets, such as in WN18RR, NELL-995-h100, and FB15K-237, an attacker has to reach a higher perturbation ratio to achieve an effect consequently making this attack weaker than the Label Perturbation. However, on the two small datasets, i.e., on UMLS and KINSHIP, the opposite is the case. The Parameter Perturbation leads to a larger performance drop with smaller perturbation rates on both datasets. %However, the overall effect of the Label Perturbation attack is higher on UMLS when a high perturbation rate is used. 
This shows that KGE models learned on the larger datasets are less susceptible to perturbations compared to the models learned on the smaller datasets. One possible explanation for such an outcome can be that the embedding space of the model learned from the larger datasets is broader and hence, a higher perturbation ratio is needed to cause significant changes on the model. On the contrary, the models learned on smaller datasets have less widened embedding spaces and even a little perturbation can lead to a large impact on the performance.

\section{Conclusion}

In this work, we have introduced non-adversarial attacks considering three attack surfaces of KGE models. %The attack approaches consider several ratios to perform perturbations on the graph, label, and parameters. 
We have performed such non-adversarial attacks on 5 state-of-the-art KGE algorithms considering 5 datasets across 3 attack surfaces, considering 8 different perturbation ratios. Our results suggest that non-adversarial attacks on different surfaces have different rates of performance degradation changes. While attacking the graph by considering lower perturbation ratios can lead to performance improvements, the same ratio can completely degrade the performance when considering the label perturbation. 

Therefore, the findings emphasize the importance of evaluating KGE models against different types of perturbations to ensure their robustness, especially if they are to be deployed in dynamic environments where the input data or the model parameters might be subject to variations. The goal would be to develop KGE models that not only perform well under ideal conditions but can also withstand and adapt to unexpected changes in their operational parameters. Potential approaches could include the development of models that inherently account for parameter variability, the use of robust optimization techniques, or the implementation of adaptive learning rates that could mitigate the impact of high perturbation ratios. Moreover, we envision future research exploring how perturbations can be leveraged to improve KGE model performance effectively. This study serves as an initial step toward a broader investigation into enhancing the robustness of KGE models.

\bibliography{references.bib}

\end{document}